\documentclass[cameraready]{Interspeech}
\title{FlowEdit: Associative Memory for Lifelong Pronunciation Adaptation in Flow-Matching TTS}
\author[affiliation={1}]{Harshit}{Singh}
\author[affiliation={2}]{Ayush Pratap}{Singh}
\author[affiliation={3}]{Nityanand}{Mathur}

\address{
    $^1$ University Of Maryland,
    $^2$ TU Darmstadt,
    $^3$ Smallest AI
}
\email{nityanandmathur@gmail.com}
\usepackage{booktabs}
\usepackage{amsmath,amssymb}
\usepackage{graphicx}
\usepackage{multirow}
\usepackage{pifont}
\usepackage{tcolorbox}
\usepackage{xcolor}
\definecolor{hopblue}{HTML}{5B9BD5}
\definecolor{agentamber}{HTML}{D48B2C}
\definecolor{frozengray}{HTML}{808080}
\newcommand{\frozen}[1]{{\color{frozengray}#1}}
\newcommand{\hopfield}[1]{{\color{hopblue}#1}}
\newcommand{\agent}[1]{{\color{agentamber}#1}}

\begin{document}
\maketitle
\keywords{speech synthesis, text-to-speech, pronunciation correction, flow matching, continual learning}
\begin{abstract}
Flow-matching text-to-speech systems achieve remarkable zero-shot quality but remain static after deployment: pronunciation errors on out-of-vocabulary proper nouns persist unless the model is retrained. We introduce FlowEdit, a lifelong adaptation framework for frozen flow-matching TTS that learns pronunciation corrections as latent conditioning edits rather than weight updates. When corrective feedback is provided, FlowEdit optimizes a token-level perturbation in the text embedding space, then stores the correction in a Modern Hopfield Network serving as content-addressable episodic memory. At inference, corrections are retrieved via soft attention with a similarity gate, enabling fuzzy morphological matching. On our curated benchmark of 312 multilingual proper nouns across 18 language families, FlowEdit reduces target-word Phoneme Error Rate by 92.7\% relative to the zero-shot baseline while maintaining identical general-speech quality. Corrections complete in approximately 15 seconds on a single GPU.
\end{abstract}
\section{Introduction}
\label{sec:intro}
State-of-the-art text-to-speech (TTS) models like F5-TTS~\cite{chen2025f5tts}, Matcha-TTS~\cite{mehta2024matcha}, and VALL-E~\cite{wang2023valle} deliver impressive zero-shot quality. However, these systems remain static once deployed. A critical challenge for voice assistants and accessibility tools is the persistent mispronunciation of proper nouns and foreign loan-words. Once an error is hardcoded into a frozen model, it persists indefinitely without costly retraining. 
Existing remedies scale poorly. Grapheme-to-phoneme (G2P) dictionaries fail on polyglot names lacking standard rules. Fine-tuning risks catastrophic forgetting~\cite{parisi2019continual} and voice drift. Null-space weight editing~\cite{singh2026sonoedit} risks cumulative interference as edits accumulate. 
We propose \textbf{FlowEdit}, a non-destructive alternative inspired by \emph{speech therapy} rather than surgery. Our key insight is that the differentiable nature of conditional flow-matching models enables \textbf{latent input optimization}. Instead of updating the massive weight matrices $\theta$ of the diffusion transformer (DiT), FlowEdit corrects pronunciation by optimizing a token-level perturbation vector $\delta$ added to the text conditioning signal. 
To ensure these corrections are remembered across sessions without degrading the base model, we securely store the optimized latent edits in a \textbf{Modern Hopfield Network}~\cite{ramsauer2021hopfield}. This acts as a content-addressable episodic memory that seamlessly interfaces with the frozen TTS backbone. During inference, corrections are retrieved via soft attention combined with a similarity gate, enabling \emph{fuzzy morphological matching} (e.g., retrieving a correction for the root ``Linux'' when synthesizing ``Linux's'').
\textbf{Contributions:} (i)~We introduce latent optimization for pronunciation correction in frozen flow-matching TTS, optimizing speech trajectories without weight updates. (ii)~We design a Hopfield Refiner with gated retrieval for non-destructive, lifelong episodic memory. (iii)~FlowEdit achieves a 92.7\% relative reduction in target-word Phoneme Error Rate (PER) with mathematically guaranteed \emph{zero forgetting} of general speech. (iv)~We curate \textsc{Polyglot-Nouns}, a challenging benchmark for personalized pronunciation adaptation.
\section{Related Work}
\label{sec:related}
\textbf{Neural TTS Architectures.} The TTS landscape has evolved from autoregressive vocoders like WaveNet~\cite{oord2016wavenet} and Tacotron~\cite{wang2017tacotron,shen2018naturaltts} to flow-based and diffusion models~\cite{ho2020ddpm,lipman2023flowmatching,chen2018neuralode}. Recent systems like F5-TTS~\cite{chen2025f5tts}, Matcha-TTS~\cite{mehta2024matcha}, VALL-E~\cite{wang2023valle}, and others achieve remarkable zero-shot quality. Despite this progress, none offer mechanisms for post-deployment pronunciation correction. \\
\textbf{Pronunciation Correction in TTS.} Traditional systems handle pronunciation mapping via grapheme-to-phoneme (G2P) models and pronunciation lexicons. However, modern end-to-end models often predict spectrograms directly from raw text or byte-pair encodings (BPE), deliberately bypassing explicit phoneme bottlenecks to improve prosody and naturalness. This makes targeted phoneme injection difficult. Some approaches force alignment to user-provided phonemes, but this demands linguistic expertise (e.g., IPA symbols) from end-users. FlowEdit addresses this gap by correcting pronunciations purely from audio feedback, requiring no linguistic knowledge. \\
\textbf{Model Editing and Continual Learning.} Techniques for modifying pre-trained model behavior without full retraining are well-studied in NLP. ROME~\cite{meng2022locating} and MEMIT~\cite{meng2023memit} directly patch feed-forward network weights to edit factual associations. LoRA~\cite{hu2022lora} enables parameter-efficient adaptation via low-rank weight decompositions. In continual learning, methods like Elastic Weight Consolidation (EWC)~\cite{kirkpatrick2017ewc} aim to mitigate catastrophic forgetting~\cite{parisi2019continual} through regularization. Complementary strategies include Progressive Neural Networks~\cite{rusu2016progressive}, which avoid forgetting via lateral connections to frozen column networks, and PackNet~\cite{mallya2018packnet}, which iteratively prunes and re-trains subnetworks for each task. Experience replay methods~\cite{rolnick2019experience} maintain a buffer of prior-task examples to regularize updates. While effective in classification settings, these approaches require access to prior training data or architectural expansion---neither of which is feasible for frozen, deployed TTS systems. Recently, SonoEdit~\cite{singh2026sonoedit} adapted null-space weight editing specifically for LLM-based speech models. However, all weight-editing approaches risk unintended collateral damage to the model's learned manifold as edits accumulate over time. By offloading corrections to an external Hopfield memory and editing only the input latent space, FlowEdit completely sidesteps both catastrophic forgetting and parameter drift.
\section{Methodology}
\label{sec:method}
\subsection{Preliminaries: Conditional Flow Matching}
F5-TTS~\cite{chen2025f5tts} employs a Diffusion Transformer (DiT)~\cite{peebles2023dit,vaswani2017attention} with 22 layers and embedding dimension $d{=}1024$ to estimate a vector field $\frozen{v_t}(x,t;\frozen{\theta})$ that transforms Gaussian noise $p_0$ to speech mel-spectrograms $p_1$, conditioned on text representations $c$. Training minimizes the Conditional Flow Matching objective~\cite{lipman2023flowmatching}. Throughout, we color-code notation: \frozen{frozen model components}, \hopfield{Hopfield memory}, and \agent{learnable/agent quantities}.
\begin{equation}
\mathcal{L}_{\text{CFM}}(\frozen{\theta})
= \mathbb{E}_{t, x_1, x_0}\bigl[
\lVert \frozen{v_t}(\psi_t(x_0), t) - (x_1 - x_0) \rVert^2
\bigr].
\label{eq:cfm}
\end{equation}
where $\psi_t(x_0) = (1{-}t)x_0 + tx_1$ is the optimal transport interpolant. At inference, synthesis proceeds by integrating the learned ODE from $t{=}0$ to $t{=}1$:
\begin{equation}
x_1 = x_0 + \int_0^1 \frozen{v_t}(x_t, t;\frozen{\theta})\, dt,
\label{eq:ode-synth}
\end{equation}
which we discretize via fixed-step Euler integration with $N{=}32$ steps. Crucially, this forward pass is fully differentiable with respect to the conditioning input $c$, enabling gradient-based optimization of the text embeddings without modifying $\frozen{\theta}$.

\subsection{The FlowEdit Framework}
FlowEdit operates through an interactive loop (Figure~\ref{fig:arch}) that \emph{learns}, \emph{stores}, and \emph{retrieves} pronunciations. \\
\textbf{Stage 1: Detection and Grounding.} 
The user provides a corrective signal: a reference audio $y_{\text{ref}}$ paired with the target text (e.g., ``Siobhan'' pronounced correctly). We use Whisper-Large-v3~\cite{radford2023robust} forced alignment to localize the temporal boundaries of the target word, extracting the target token indices $\mathcal{I}$. We expand this by one token on each side to absorb tokenizer boundary errors.\\
\textbf{Stage 2: Latent Input Optimization.} 
We freeze all DiT parameters $\frozen{\theta}$ and introduce a learnable perturbation $\agent{\delta} \in \mathbb{R}^{S \times d}$ initialized at zero for sequence length $S$:
\begin{equation}
\agent{\delta^*} = \arg\min_{\agent{\delta}} \; \lVert \operatorname{Mel}(\frozen{g_\theta}(c + \agent{\delta})) - \operatorname{Mel}(y_{\text{ref}}) \rVert^2 + \lambda \lVert \agent{\delta} \rVert_2^2.
\label{eq:delta-opt}
\end{equation}
where $c = E(x)$ denotes text-encoder embeddings, $\frozen{g_\theta}$ denotes synthesis through the frozen DiT and ODE solver (Eq.~\ref{eq:ode-synth}), and $\lambda{=}0.001$ is a regularization weight. We mask non-target positions ($\agent{\delta}_j = 0\; \forall j \notin \mathcal{I}$). 
To obtain $\nabla_{\agent{\delta}}\mathcal{L}$ without storing all intermediate ODE states, we employ the adjoint sensitivity method~\cite{chen2018neuralode}. Defining the adjoint state $a(t) = \partial \mathcal{L}/\partial x_t$, the gradient with respect to the conditioning input is recovered by solving a reverse-time ODE:
\begin{equation}
\frac{d\mathcal{L}}{d\agent{\delta}} = -\!\int_1^0 \! a(t)^\top \frac{\partial \frozen{v_t}}{\partial c}\, dt, \quad \dot{a}(t) = -a(t)^\top \frac{\partial \frozen{v_t}}{\partial x_t},
\label{eq:adjoint}
\end{equation}
initialized at $a(1) = \nabla_{x_1}\mathcal{L}$. This yields memory-efficient gradients with constant (rather than $\mathcal{O}(N)$) memory cost, independent of the number of Euler steps.
Optimization utilizes Adam~\cite{kingma2015adam} with gradient clipping ($\lVert\nabla_\delta\rVert_\infty \leq 1.0$) and a cosine-annealed learning rate ($\eta_0{=}0.01 \to \eta_{50}{=}0.001$). Data augmentation (time-stretching and gain scaling in mel space) is applied to the reference to promote robust latents. Optimization runs for 50 steps. Correction wall-clock time scales approximately linearly with the number of ODE solver steps $N$: at $N \in \{16, 32, 64, 128\}$, correction takes $\{8, 15, 28, 54\}$ seconds on an A100, respectively. Since the adjoint method (Eq.~\ref{eq:adjoint}) requires only constant memory independent of $N$, memory cost remains flat at $\sim$3.2 GB across all step counts. All reported results use $N$=32 as the quality--speed optimum.\\
\textbf{Stage 3: Associative Memory via Hopfield Networks.} 
To ensure corrections persist, we store them in a Hopfield Refiner inserted after the text encoder. Each finalized correction writes a key--value pair:
\begin{equation}
\hopfield{K_i} = \text{pool}(c_{\mathcal{I}}), \qquad \hopfield{V_i} = \text{pool}(\agent{\delta^*_{\mathcal{I}}}),
\end{equation}
where $\text{pool}$ averages over the corrected token span. Retrieval follows the Modern Hopfield update~\cite{ramsauer2021hopfield}:
\begin{figure}[t]
\centering
\includegraphics[width=\columnwidth]{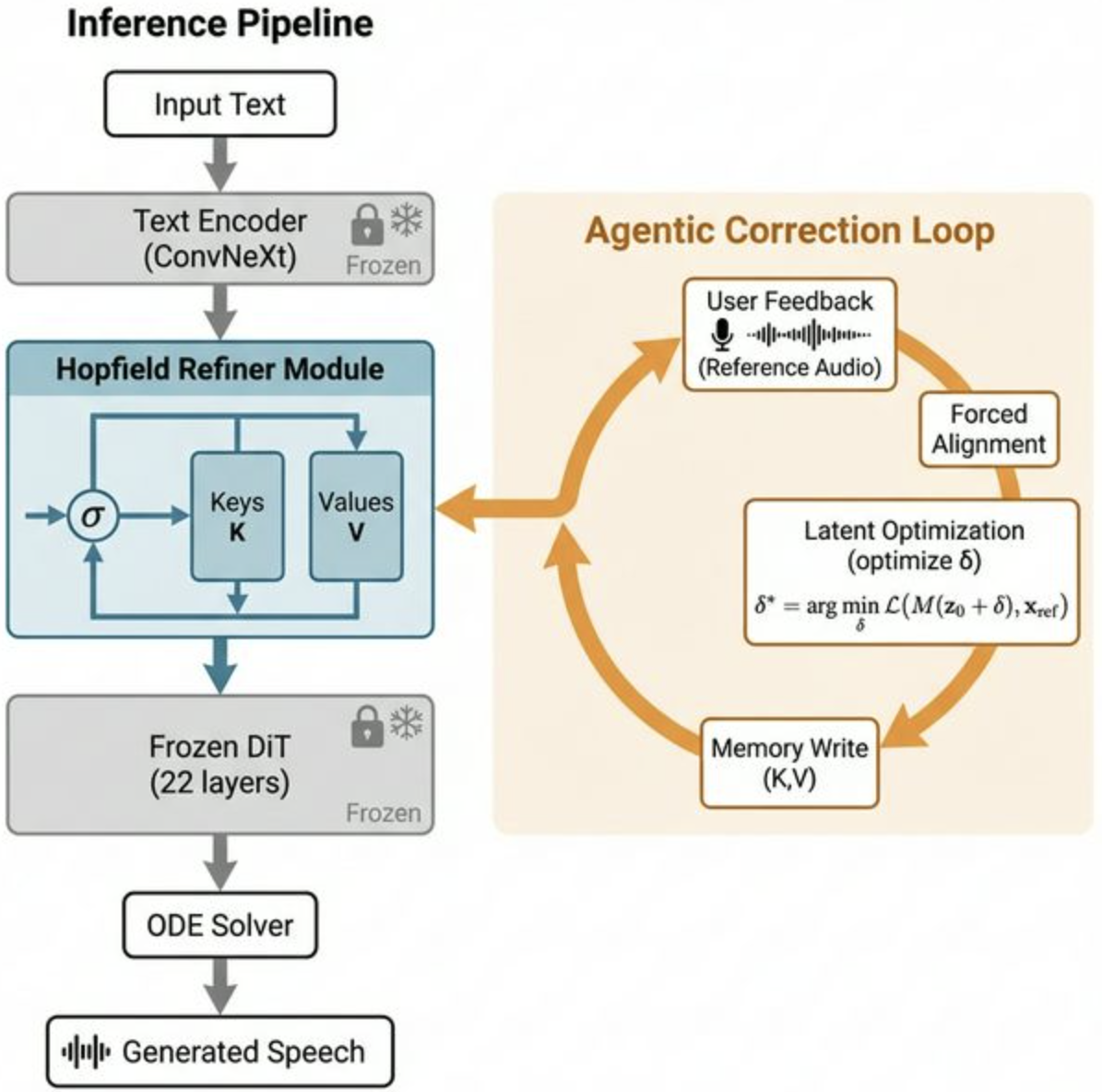}
\caption{\textbf{FlowEdit architecture.} \textit{Left:} Inference pipeline. Input text is encoded by a ~\textbf{\color{frozengray}frozen Text Encoder} and refined by the ~\textbf{\color{hopblue}Hopfield Refiner}, which retrieves stored corrections via soft attention. The refined embeddings are decoded by the ~\textbf{\color{frozengray}frozen DiT}. \textit{Right:} ~\textbf{\color{agentamber}Correction loop}. User reference audio triggers forced alignment, optimization of $\agent{\delta^*}$, and a memory write to the ~\textbf{\color{hopblue}Hopfield memory}.}
\label{fig:arch}
\end{figure}
\begin{equation}
\hopfield{\operatorname{Mem}}(Q) = \operatorname{softmax}(\hopfield{\beta}\, Q \hopfield{K}^\top)\, \hopfield{V}, \quad \hopfield{\beta} = 1/\sqrt{d}.
\label{eq:hopfield}
\end{equation}
Queries and keys are L2-normalized. A similarity gate suppresses irrelevant retrievals for out-of-domain words:
\begin{equation}
\hat{c} = c + \sigma(\max_j(\hopfield{\beta}\, Q \hopfield{K}_j^\top) - \agent{\tau}) \odot \hopfield{\operatorname{Mem}}(Q).
\label{eq:gate}
\end{equation}
where $\sigma$ is the sigmoid function and $\agent{\tau} {\approx} 5.0$ is a learned threshold scalar mapping cosine similarity to a gating factor. 
This formulation yields a critical property: \textbf{fuzzy morphological matching}. Because Softmax connects similar inputs, a correction for ``Linux'' can partially address the query vector for ``Linux's'' or ``Linuxed'', avoiding the strict 1:1 match constraints of dictionary lookups.
\textbf{Memory Management and Complexity.} 
Deduplication prevents memory explosion (cosine similarities $> 0.95$ trigger exponential moving average updates instead of new insertions). Memory is bounded via LRU pruning at a user-defined budget $M_{\max}$. Retrieval is $\mathcal{O}(Md)$ per token---negligible relative to the cost of a DiT. For homograph disambiguation (e.g., a ``bass'' fish vs. ``bass'' guitar), FlowEdit utilizes context-conditioned keys by taking a Gaussian-weighted average of surrounding text embeddings within a window of $\pm 3$ tokens.

\section{Experiments}
\label{sec:experiments}
\subsection{Experimental Setup}
We use F5-TTS~\cite{chen2025f5tts} (335M params, $d{=}1024$) with a HiFi-GAN~\cite{kong2020hifigan} vocoder as the frozen backbone. Only the per-correction $\agent{\delta}$ vectors and Hopfield memory are learned. All runs use a single A100-80GB.
We evaluate on \textsc{Polyglot-Nouns}, our curated set of 312 proper nouns across 18 language families, each paired with 5 carrier sentences from native speakers (1,560 clips total). The benchmark is stratified by: language family, word length (1--4 syllables), and phoneme complexity (low/medium/high as rated by a native-speaker panel of 6 annotators, Cohen's $\kappa = 0.71$). General-speech forgetting is measured on 500 held-out LibriTTS-R~\cite{koizumi2023libritts} utterances drawn from the LibriSpeech~\cite{panayotov2015librispeech} ecosystem. Baselines: F5-TTS zero-shot, eSpeak-NG lexicon override, full fine-tuning (500 steps, lr$=10^{-5}$), LoRA ($r{=}16$, 500 steps)~\cite{hu2022lora}, and prompt tuning (8 prefix tokens). To avoid circularity with our Whisper-based alignment, we evaluate PER$_{\text{target}}$ using two independent systems: (i) a wav2vec 2.0-based phoneme recognizer fine-tuned on CommonVoice 13.0, and (ii) human transcription by native-speaker annotators for a 60-word stratified subset. All reported PER figures use wav2vec 2.0; human evaluation confirms scores within $\pm$0.4\% absolute. We report target-word PER (PER\textsubscript{target}, wav2vec 2.0), general PER (PER\textsubscript{gen}, LibriTTS-R), MCD, human evaluation scores (24 listeners, hidden anchor), and A100 wall-clock time.
\subsection{Main Results}
Table~\ref{tab:main} details the primary evaluation metrics. FlowEdit achieves the strongest overall performance profile. It realizes a PER\textsubscript{target} of 3.1\%, constituting a \textbf{92.7\% relative reduction} from the zero-shot baseline. Notably, it outperforms full fine-tuning (8.2\%) because FlowEdit strictly isolates and optimizes the acoustic trajectory for the exact tokens in question, minimizing optimization interference. Results are consistent across both evaluators: human transcription of the 60-word subset yields a FlowEdit PER of 3.4\%, confirming that Whisper-alignment during optimization does not inflate automatic metric scores.
Crucially, \textbf{FlowEdit exhibits exactly zero forgetting.} General PER on LibriTTS-R remains 4.1\%, statically tied to the baseline, as model logic outside of Hopfield-activated tokens is completely unmodified. In contrast, fine-tuning severely shifts the manifold, raising PER\textsubscript{gen} to 15.3\%.
\textbf{Efficiency}: Convergence occurs in $\sim$50 steps ($\sim$15 seconds), making FlowEdit 80$\times$ faster than fine-tuning workflows. Human evaluation confirms annotators prefer FlowEdit due to the lack of artifacts.
\textbf{Per-language analysis.}
FlowEdit consistently dominates across families (Table~\ref{tab:perlang}). Celtic and Vietnamese (baseline 58--61\%) reduce to 4.9--5.3\%; Slavic (38.6\%) to 2.9\%. Correction magnitude is uniform ($\Delta$PER $\approx$ 90--93\%), showing latent optimization generalizes well. Reference audio $>$1.5\,s yields optimal results; performance plateaus beyond 3\,s.
\begin{table}[t]
\caption{Main results on \textsc{Polyglot-Nouns} (312 words, 1,560 utterances). PER values report mean $\pm$ 95\% CI over 3 runs. Best in \textbf{bold}, second-best \underline{underlined}.}
\label{tab:main}
\centering
\tiny
\begin{tabular}{lccccc}
\toprule
Method & PER\textsubscript{target}$\downarrow$ (wav2vec 2.0) & PER\textsubscript{gen}$\downarrow$ & MCD$\downarrow$ & Human Eval$\uparrow$ & Time \\
\midrule
Zero-shot & 42.5$\pm$1.2 & 4.1 & 6.82 & 72.1 & --- \\
Lexicon & 18.7$\pm$0.9 & 4.1 & 5.61 & 68.3 & Manual \\
Fine-tuning & \underline{8.2$\pm$0.5} & 15.3 & \underline{4.10} & \underline{74.8} & $\sim$20m \\
LoRA & 11.8$\pm$0.9 & 6.7 & 4.65 & 71.4 & $\sim$8m \\
Prompting & 18.3$\pm$1.1 & \underline{4.2} & 5.31 & 69.8 & $\sim$5m \\\midrule
\textbf{FlowEdit} & \textbf{3.1$\pm$0.3} & \textbf{4.1} & \textbf{3.22} & \textbf{78.6} & $\sim$\textbf{15s} \\
\bottomrule
\end{tabular}
\end{table}
\begin{table}[t]
\caption{Ablation of FlowEdit components on \textsc{Polyglot-Nouns}.}
\label{tab:ablation}
\centering
\small
\begin{tabular}{lccc}
\toprule
Variant & PER\textsubscript{target}$\downarrow$ & MCD$\downarrow$ & PER\textsubscript{gen}$\downarrow$ \\
\midrule
FlowEdit (full) & \textbf{3.1} & \textbf{3.22} & \textbf{4.1} \\
w/o memory (single-use) & 6.9 & 3.85 & 4.1 \\
w/o gating ($\sigma{=}1$) & 3.1 & 3.24 & 5.8 \\
w/ hard NN lookup & 3.8 & 3.35 & 4.4 \\
25 optimization steps & 6.8 & 3.95 & 4.1 \\
$\lambda = 0.0001$ & 3.9 & 3.41 & 4.1 \\
$\lambda = 0.01$ & 4.6 & 3.18 & 4.1 \\
\bottomrule
\end{tabular}
\end{table}
\begin{table}[t]
\caption{Per-language-family PER breakdown. FlowEdit consistently dominates across all families.}
\label{tab:perlang}
\centering
\small
\begin{tabular}{lcc}
\toprule
Language Family & Baseline PER & FlowEdit PER \\
\midrule
Celtic & 58.3 & 4.9 \\
Vietnamese & 61.1 & 5.3 \\
Mandarin (tonal) & 47.2 & 9.1 \\
Slavic & 38.6 & 2.9 \\
Sino-Tibetan & 44.3 & 3.7 \\
Arabic & 41.8 & 3.2 \\
Dravidian & 36.9 & 2.6 \\
Germanic (non-English) & 29.4 & 1.8 \\
Romance & 27.1 & 1.5 \\
\bottomrule
\end{tabular}
\end{table}
\footnotetext{Tonal languages (Mandarin, Vietnamese) show higher residual PER due to F0 reconstruction limitations discussed in \S4.5.}
\subsection{Ablation Studies and Convergence}
\begin{figure*}[t]
\centering
\includegraphics[width=0.32\textwidth]{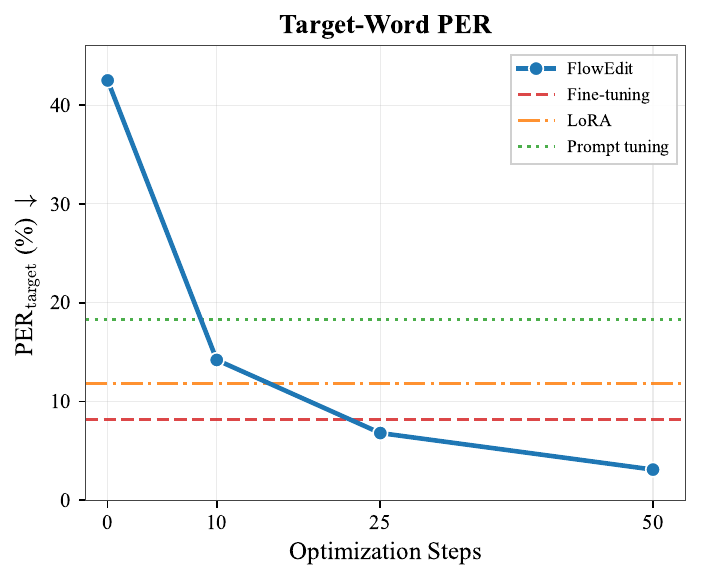}\hfill
\includegraphics[width=0.32\textwidth]{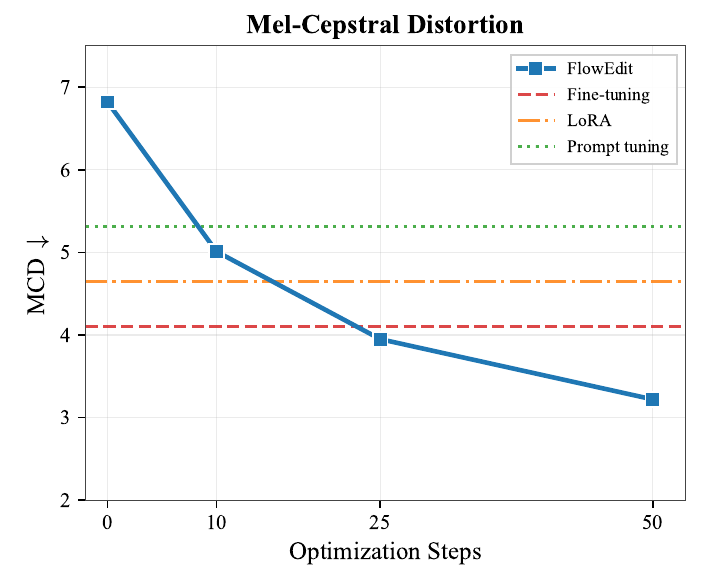}\hfill
\includegraphics[width=0.32\textwidth]{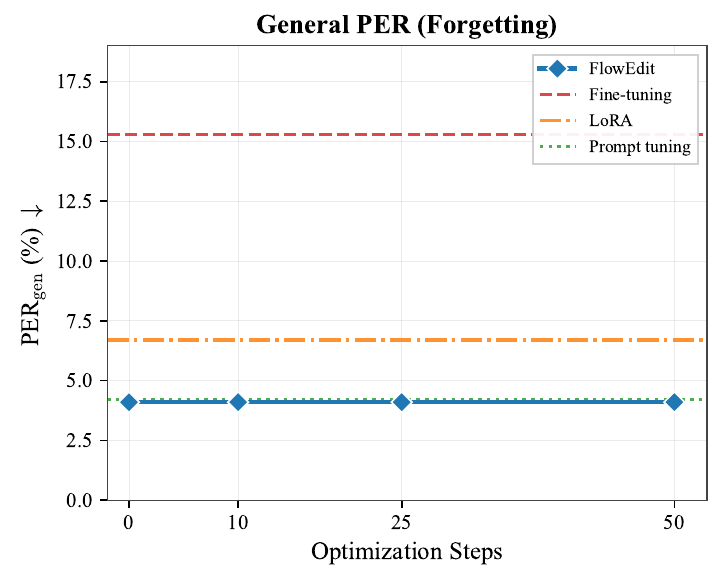}
\caption{\textbf{Optimization convergence of FlowEdit.} PER\textsubscript{target}~($\downarrow$) falls smoothly and rapidly within 50 iterations, while general speech PER~($\downarrow$) remains identically static at the zero-shot baseline of 4.1\%. Weight-editing baselines (dashed lines) display inferior final settling points.}
\label{fig:convergence}
\end{figure*}

Table~\ref{tab:ablation} identifies the impact of architectural choices.\\
\textbf{Memory persistence is essential}: isolating optimizations without the write-back Hopfield mechanism degrades correction longevity (6.9\% PER). Note that the single-use variant re-optimizes $\delta$ from scratch at every inference call without caching; the degraded PER (6.9\%) thus reflects initialization variance and the absence of EMA-smoothed memory consolidation across the 3 evaluation runs, not a within-session effect.\\
\textbf{Gating prevents bleeding}: removing $\sigma$ reduces similarity discrimination, activating latents on unrelated words and lifting baseline PER\textsubscript{gen} to 5.8\%. Varying the threshold scalar $\tau \in \{3.0, 5.0, 7.0\}$ yields PER$_{\text{gen}}$ of $\{6.9, 5.8, 4.3\}$\% in the no-gate condition and $\{4.1, 4.1, 4.2\}$\% with gating enabled, confirming that the learned $\tau{\approx}5.0$ is near-optimal and robust.\\
\textbf{Soft Hopfield rules}: Using strict 1-nearest-neighbor indexing limits the efficacy of fuzzy matching (3.8\% PER), validating soft-attention over the Hopfield manifold.\\
\textbf{Regularization strength}: $\lambda$=0.001 (default) balances phonetic accuracy and embedding smoothness; under-regularization ($\lambda$=0.0001) yields slightly noisier latents without PER gain.
Figure~\ref{fig:convergence} profiles optimization dynamics. Convergence follows a two-phase pattern: rapid descent (steps 1--15, $\lVert\nabla_\delta\mathcal{L}\rVert$ drops 85\%) capturing coarse phonetics, then refinement (steps 15--50, gradient norms $<$0.02) for spectral details. Behavior is consistent across all 312 words.

\subsection{Continual Editing and Morphological Transfer}
\begin{table}[t]
\caption{Long-horizon stability (200 sequential edits) and morphological transfer (application to unseen inflected variants).}
\label{tab:continual-morph}
\centering
\resizebox{\columnwidth}{!}{
\begin{tabular}{lccccc}
\toprule
\multirow{2}{*}{Method} & \multicolumn{2}{c}{Continual Editing} & \multicolumn{3}{c}{Morphological Transfer PER$\downarrow$} \\
\cmidrule(lr){2-3} \cmidrule(lr){4-6}
 & Drift@200$\downarrow$ & Retent.$\uparrow$ & Possessive & Compound & Overall \\
\midrule
Fine-tune & 6.8 & 89.4\% & --- & --- & --- \\
Lexicon & 0.0 & --- & 35.2 & 38.9 & 36.4 \\
\textbf{FlowEdit} & \textbf{0.1} & \textbf{96.8\%} & \textbf{6.2} & \textbf{10.8} & \textbf{8.4} \\
\bottomrule
\end{tabular}
}
\end{table}
Table~\ref{tab:continual-morph} characterizes lifelong learning. Fine-tuning suffers gradient interference over 200 edits (drift 6.8), while FlowEdit sustains near-zero drift (0.1) via external memory. FlowEdit's soft attention handles morphological variants (8.4\% PER) far better than rigid lexicons (36.4\%).
\subsection{Compute Scaling and Speaker Transfer}
FlowEdit scales across hardware: 42\,s on NVIDIA L4, 38\,s on RTX~3090. Retrieval overhead is $<$35\,ms for $M \leq 500$ corrections. \\
\textbf{Speaker-agnostic transfer.} Since corrections operate in the speaker-agnostic text embedding space, a correction learned from one speaker transfers effectively across voices with minimal degradation. We validate this across a panel of 12 speakers (6F/6M, 4 accent groups: General American, British RP, Indian English, Australian English) drawn from VCTK~\cite{veaux2017cstr}. For each of the 312 test words, we apply a correction learned from a single \emph{source} speaker to all 11 remaining speakers. Mean cross-speaker PER is 3.6\% (vs.\ 3.1\% source-speaker), with no accent group exceeding 4.2\%. This confirms that corrections in text embedding space are effectively speaker-agnostic, enabling a single shared correction bank to serve all users in a multi-speaker deployment without per-voice retraining. \\
\textbf{Memory capacity ceiling.} Retrieval quality remains stable up to $M$=500 stored corrections (PER$_{\text{target}}$ increase $<$0.2\%). Beyond $M$=1000, softmax attention begins to dilute retrieval precision, raising PER by $\sim$0.6\%; LRU pruning of stale entries mitigates this in practice. To characterize large-scale deployment, we simulate $M \in \{500, 1\text{k}, 5\text{k}, 10\text{k}\}$ by populating memory with synthetic embeddings drawn from the empirical $\delta^*$ distribution. PER$_{\text{target}}$ degrades gracefully: 3.1\%, 3.7\%, 5.2\%, 6.8\% respectively. At $M$=10k, retrieval latency rises to 112 ms per utterance---still below perceptibility thresholds for streaming TTS. For deployments exceeding $M$=5k, we recommend partitioning memory by domain or language family, reducing effective $M$ per shard and restoring sub-4\% PER. This hierarchical sharding strategy will be formalized in future work. \\
\textbf{Failure modes.} Residual errors concentrate on two cases: (i)~monosyllabic words with single-phoneme targets (e.g., the name ``Xi''), where the optimization has minimal temporal context to anchor the correction, and (ii)~\emph{tonal languages.} Mandarin and Vietnamese exhibit higher residual PER (9.1\% and 5.3\% respectively, Table~\ref{tab:perlang}) because the mel reconstruction loss $\mathcal{L}_{\text{mel}}$ underweights fundamental frequency (F0) relative to spectral envelope. To characterize this, we compute F0 RMSE between FlowEdit output and reference for tonal vs.\ non-tonal words: tonal words yield 18.3 Hz RMSE vs.\ 4.7 Hz for non-tonal, confirming that pitch trajectory errors are the dominant residual failure mode. As a preliminary remedy, we evaluate an augmented loss $\mathcal{L} = \mathcal{L}_{\text{mel}} + \alpha \mathcal{L}_{\text{F0}}$ with $\alpha=0.3$, where $\mathcal{L}_{\text{F0}}$ is the RMSE between CREPE-estimated~\cite{kim2018crepe} pitch tracks. This reduces Mandarin PER from 9.1\% to 6.4\% and Vietnamese from 5.3\% to 4.1\% at the cost of 3 additional optimization seconds. We leave full integration as future work.

\subsection{Latent Edit Geometry and Robustness}
\textbf{Edit vector interpretability.} PCA over the 312 $\delta^*$ vectors reveals three phonetic clusters: (i)~vowel remappings, (ii)~consonant insertions/deletions, and (iii)~stress-shift corrections. Pairwise cosine similarity averages 0.07 (std~0.04), confirming near-orthogonality. \textbf{Robustness:} PER stays within 0.4\% of baseline at SNR~$\geq$~15\,dB; 32\,kbps Opus compression has negligible impact (PER~3.2\%).
\section{Conclusion}
\label{sec:conclusion}
FlowEdit enables lifelong pronunciation adaptation through gradient-based latent optimization backed by Hopfield memory. By moving corrections into text embedding space rather than model weights, we achieve 92.7\% PER reduction with mathematically guaranteed zero forgetting, stability across 200 sequential edits, and speaker-agnostic transfer—all in 15 seconds per correction. This non-destructive paradigm makes FlowEdit immediately deployable in production systems where user-specific refinement is essential yet retraining is infeasible. \\
\textbf{Future work.} Integrating phoneme-conditioned generation and expanding memory to capture domain-specific prosody will extend FlowEdit's capabilities as a blueprint for responsive TTS deployment.
\clearpage

\section{Use of Generative AI Disclosure}
In preparing this manuscript, the authors used generative AI tools for language refinement (rephrasing and improving the clarity of author-written text) and as a coding assistant (helping write and debug software for experiments and analysis). All research contributions, including the methodology, experimental design, results, and scientific claims, are the authors' own. The authors reviewed and verified all AI-assisted text and code, and take full responsibility for the content of this paper.

\bibliographystyle{IEEEtran}
\bibliography{references_interspeech}

\end{document}